# PESAO: Psychophysical Experimental Setup for Active Observers

## Tech-Report

Markus D. Solbach John K. Tsotsos

{solbach, tsotsos} @eecs.yorku.ca

Department of Electrical Engineering and Computer Science
York University, Canada

# Contents

# Table of Figures

# Introduction

Most past and present research in computer vision involves passively observed data. Humans, however, are active observers outside the lab; they explore, search, select what and how to look. Nonetheless, how exactly active observation occurs in humans so that it can inform the design of active computer vision systems is an open problem.

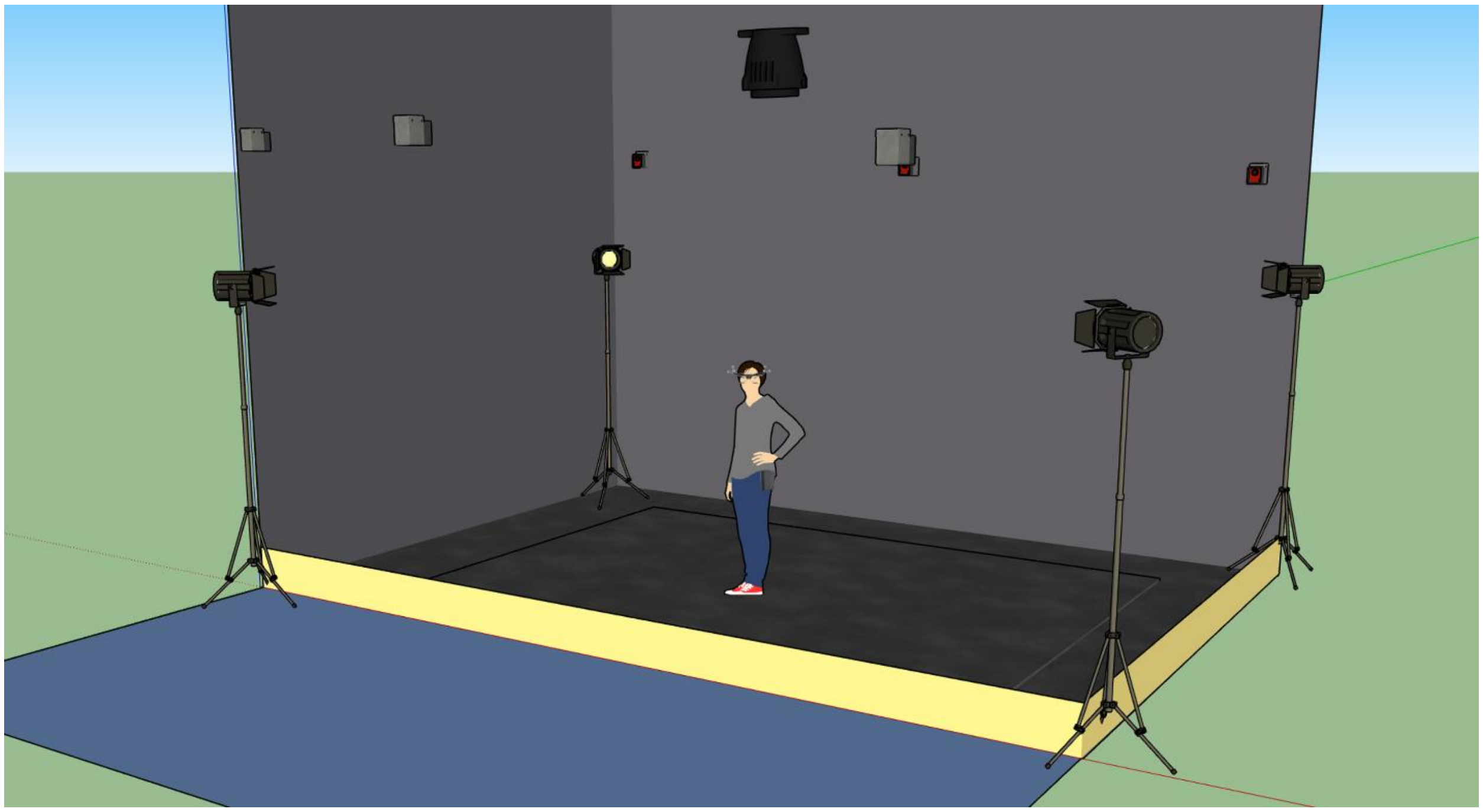

*Figure 1 Sketch of PESAO.*

PESAO is designed for investigating active, visual observation in a 3D world. The goal was to build an experimental setup for various active perception tasks with human subjects (active observers) in mind that is capable of tracking the head and gaze.

While many studies explore human performances, usually, they use line drawings portrayed in 2D, and no active observer is involved. PESAO allows us to bring many studies to the three-dimensional world, even involving active observers. In our instantiation, it spans an area of 400cm x 300cm and can track active observers at a frequency of 120Hz.

Furthermore, PESAO provides tracking and recording of 6D head motion, gaze, eye movement-type, first-person video, head-mounted IMU sensor, birds-eye video, and experimenter notes. All are synchronized at microsecond resolution.

In the next Sections, we walk through all steps needed to build your own PESAO. We describe the hardware that we have used, how to set it up, and describe PESAOlib, which is the accompanying software of PESAO. It is used to design and run an experiment and also to synchronize and analyze the data. PESAOlib is an opensource implementation developed in Python and C++. It provides basic functionalities and is extendable. Figure 1 illustrates a sketch of PESAO.

In short, PESAO is capable of:

- Head tracking
- Gaze Tracking
- Tracking of Objects
- Controlled Lighting
- Data synchronization
- Data analysis

The project page can be found under http://data.nvision2.eecs.yorku.ca/PESAO/

# Hardware

In this Section, we describe the hardware that we have used to build PESAO. The core hardware pieces are a motion tracking system and an eye-tracking system. Both will be discussed in the following subsections. We will also discuss the glasses' tracking body, which was custom made to track the glasses with the motion tracking system, object tracking bodies, our light-setup and give details on the hardware specifications.

Besides a motion tracking and an eye-tracking system, PESAO requires a workstation computer to run PESAOlib.

Based on the hardware configuration, the workstation computer should fulfill the following minimum requirements:

- Windows 10 (requirement of *Motive* motion capture software)
- > Intel i7-7700k, Ryzen 7 2700x or comparable
- > 8GB RAM
- > 128GB SSD storage
- NVIDIA Quadro, > 2GB VRAM, released 2015 or later (requirement of *Tobii Pro Lab* software)

## Motion Tracking System

In this subsection, we present the motion tracking system we have used to built PESAO. A range of products exists that is suitable; ultimately, our setup uses a system by OptiTrack.

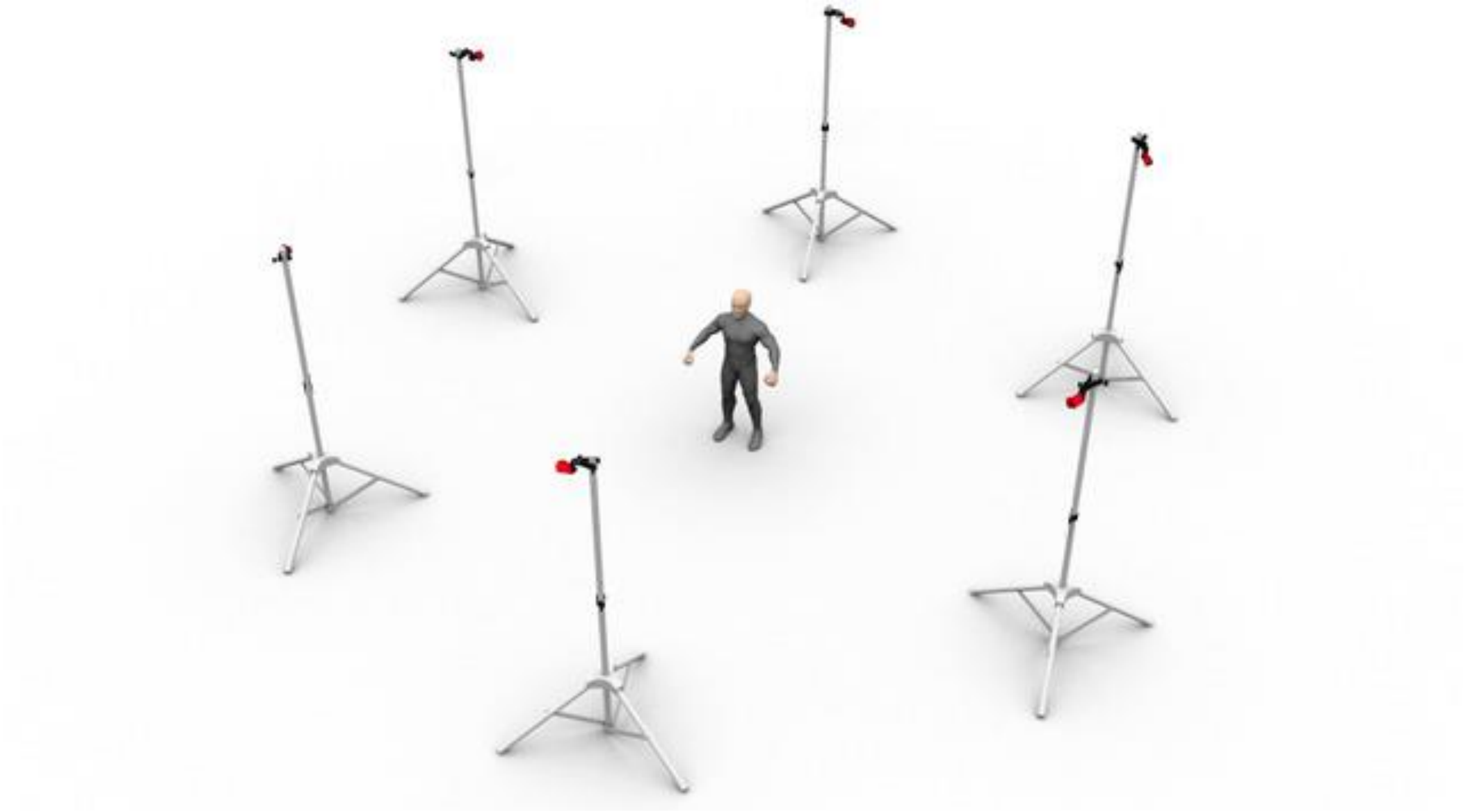

*Figure 2 Motion tracking system.*

*Courtesy of NaturalPoint, Inc., accessed 15 September, 2020, https://optitrack.com/public/images/volume12CamStand.jpg*

For our implementation of PESAO, we chose the *Robotics Package* with six *Flex 13* cameras. With this, up to ten objects can be tracked in a 4m x 4m x 2m volume at 120Hz. This set comes with the most necessary accessories. However, we purchased an additional set of 30 *M4 Markers,* and six 10ft camera stands together with six 3-way head clamps. Figure 2 illustrates *Flex 13* cameras mounted on stands.

## Eye-Tracking System

For PESAO, we chose the *Tobii Pro Glasses 2*.

They are capable of recording precise gaze information at either 50 or 100Hz (dependent on the model), first-person video, and motion information (accelerometer and gyroscope). Furthermore, for a more straightforward analysis, we also used the *Tobii Pro Lab* software to export gaze information and detect eye events. For PESAOlib to be fully functional, it requires this data to synchronize the gaze information with the remaining data.

Useful to include more subjects in the study and to serve subjects with vision impairment, we purchased the *prescription lenses package* from Tobii. This package contains corrective snap-on lenses to support subjects with either short- or long-sightedness.

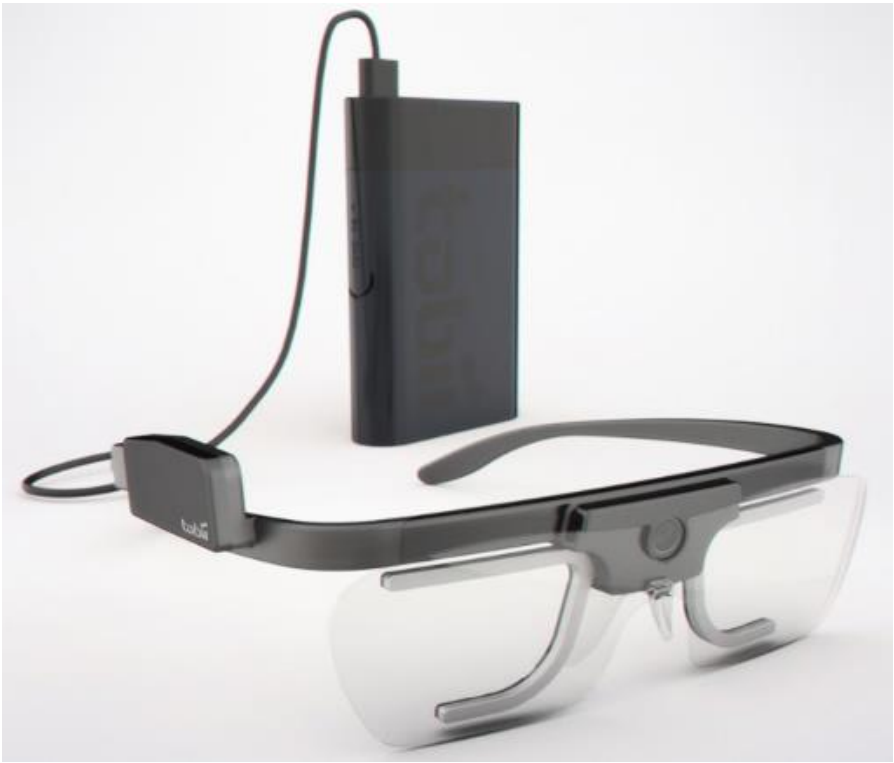

*Figure 3 Tobii Pro Glasses 2 Eye Tracking System by Tobii.*

*Courtesy of Tobii AB, accessed 15 September, 2020, https://www.tobiipro.com/product-listing/tobii-pro-glasses-2/*

## Glasses Tracking Body

In order to integrate the eye-tracking data with the motion-tracking data, we developed a custom tracking body for the Tobii Pro Glasses 2. The tracking body is a snap-on solution that works with most hairstyles and head-decorations. We put a focus on reliability, durability and modularity in order to never lose a subject to failing hardware or lose their data. The tracking body has a modular design for a piece-wise exchange of broken elements. See Figure 4 *Exploded View Drawing*.

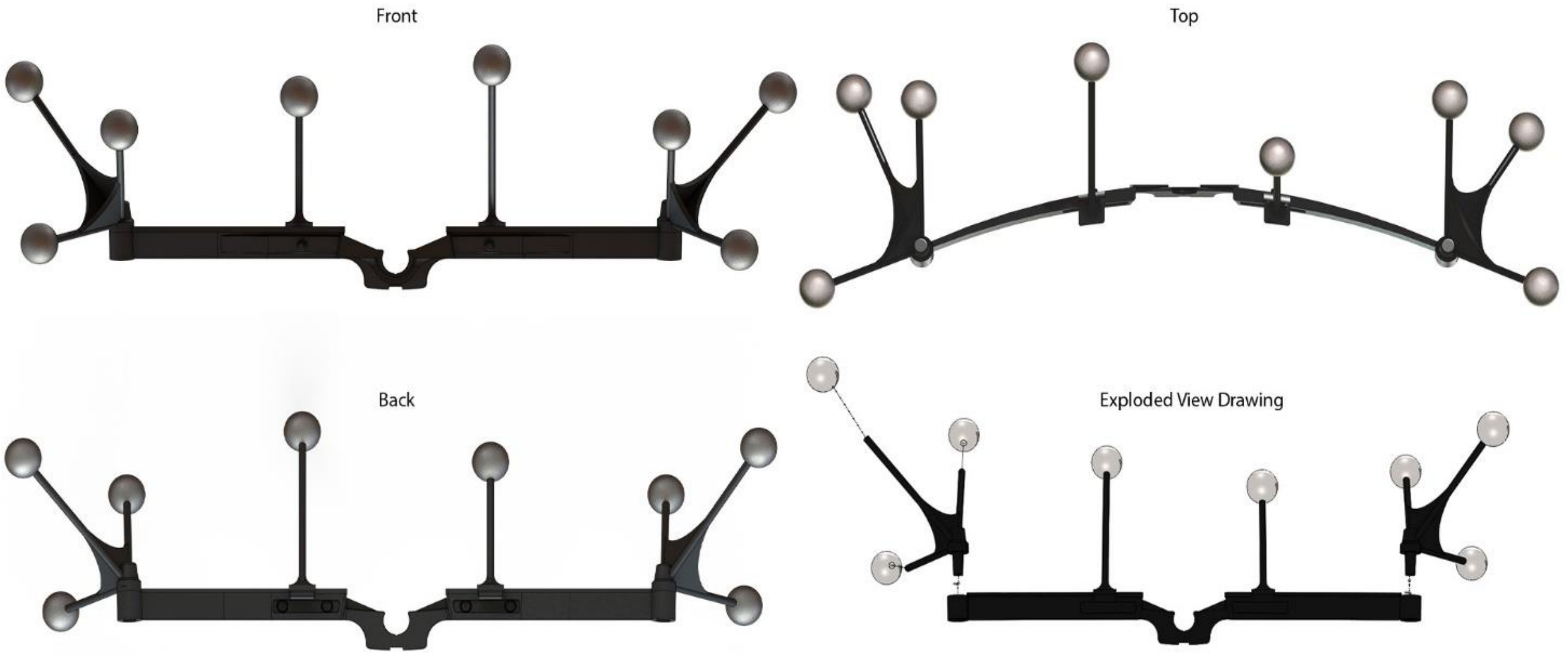

*Figure 4 Custom Tracking Body for Tobii Pro Glasses 2.*

The tracking body is equipped with standard *M4 Markers* (available through OptiTrack). If it is required to use prescription lenses with the system, it is necessary to attach four magnets to the tracking body (Ø4mm). The 3D printable file for the tracking body and print settings can be found in the README file on the project page.

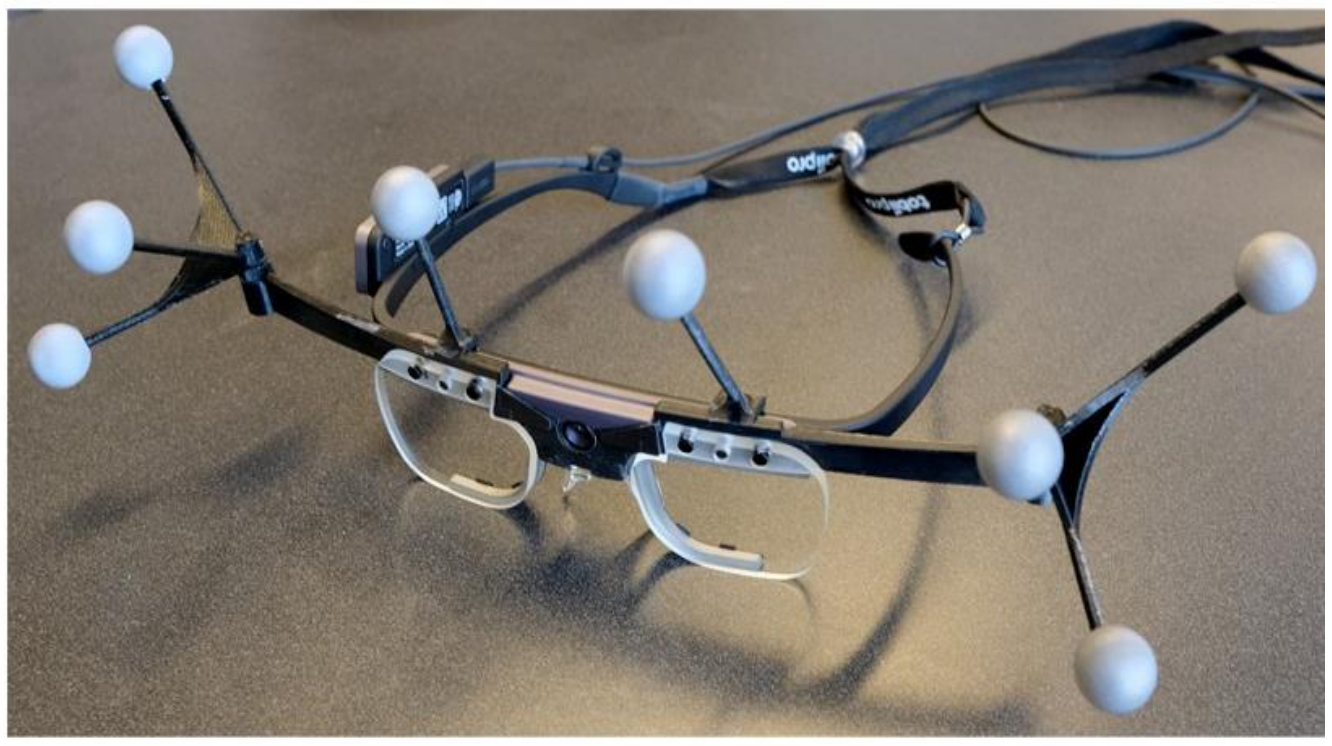

*Figure 5 Assembled tracking body.*

An assembled tracking body mounted on *Tobii Pro Glasses 2* can be seen in Figure 5. The tracking body does not change the tracking capabilities of the glasses nor obstruct the field of view of the Subject. However, it is necessary to set up and calibrate the tracking body in the *Motive* motion tracking software. As a result, PESAO records precise six-degree-of-freedom tracking information of the tracking body.

While conducting experiments, the visibility of the first-person camera of the glasses positioned between the eyes of the Subject should be checked. Subjects tend to push the glasses up their nose by pressing against the bridge of the glasses where the camera is positioned, hence smearing the camera lens. The glasses and camera lens should be regularly cleaned.

## Object Tracking Body

For several visual perception experiments, it is also of interest to track the position of the object under investigation.

We have designed two custom object tracking bodies that can be used to track objects in 3D space. The bodies can be either mounted directly on the object or on a rod with a 2.54cm diameter. The objects are designed for tracking reliability and ease of use.

Similarly to the glasses tracking body, the object tracking bodies are available as a 3D printable file. For further information on print settings, please review the README file on the project page. The tracking bodies have to be calibrated in the *Motive* motion tracking software in order to be recognized by PESAO.

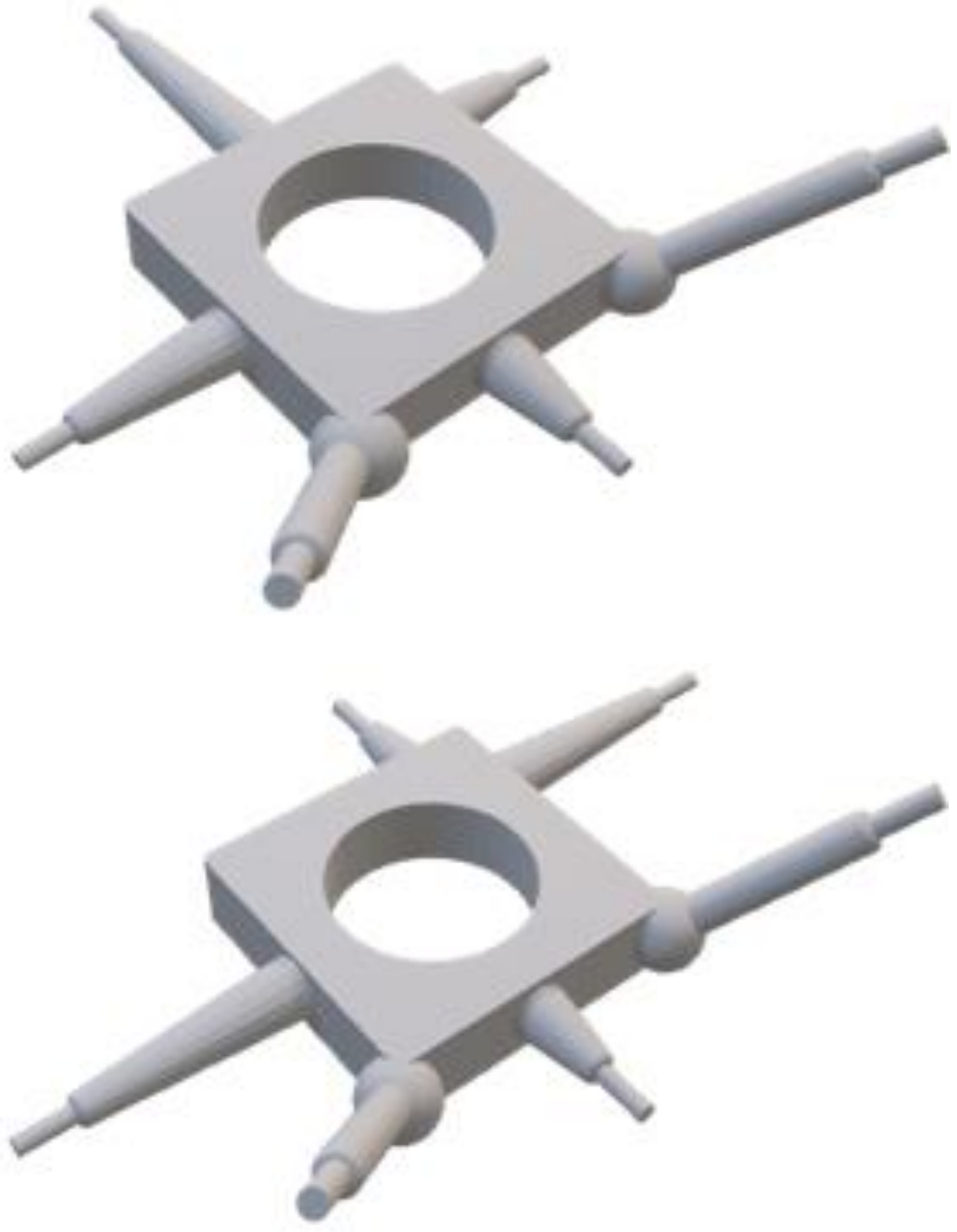

*Figure 6 Object Tracking Body one (top) and two (bottom).*

## Light

Our visual system cannot function without light; hence to study the effect of light, PESAO offers controllable light settings and light measuring capabilities.

Figure 7 shows a light source as used with PESAO. We have used 660 LED Video light-panels from Neewer. They are offered with stands. An essential feature of these lights is their capability to dim and change the colour temperature.

For our tracking area of 400 x 300cm, we have used five light sources, one in each corner and one above functioning as a ceiling light. See Figure 1.

*Note:* To eliminate shadows and to provide more homogeneous lighting, the height of the four corner lights' should be slightly below the height of the object of interest.

The light panels provide colour temperatures from 3200 – 5600K and lumen of up to 7300 Lux/m.

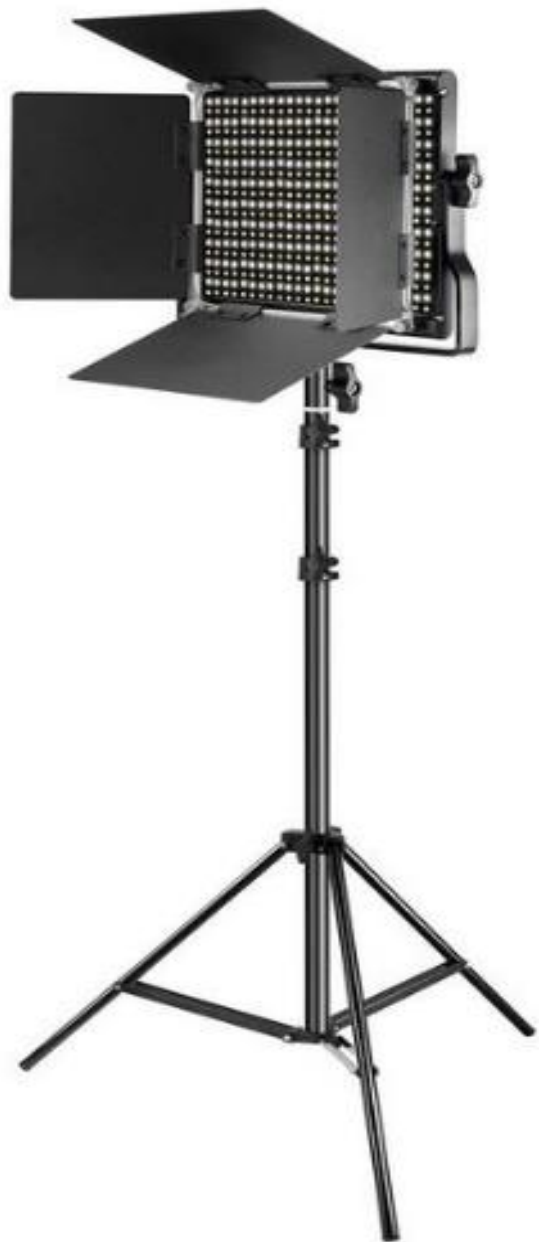

*Figure 7 LED light panel.*

*Courtesy of Neewer®, accessed 15 September, 2020, https://cdn.shopify.com/s/files/1/0225/2960/5706/products/90089829_800x.jpg*

*[Feature Outlook. Planned to be part of the next software release.]* In order to measure the actual light intensity in the scene or on the object, PESAO is designed to integrate Yocto-Light-V2 by Yoctopuce. The Yocto-Light-V2 device is a USB light sensor (lux meter) that measures light of up to 65,000 lux. A photo of the light sensor can be seen in Figure 8.

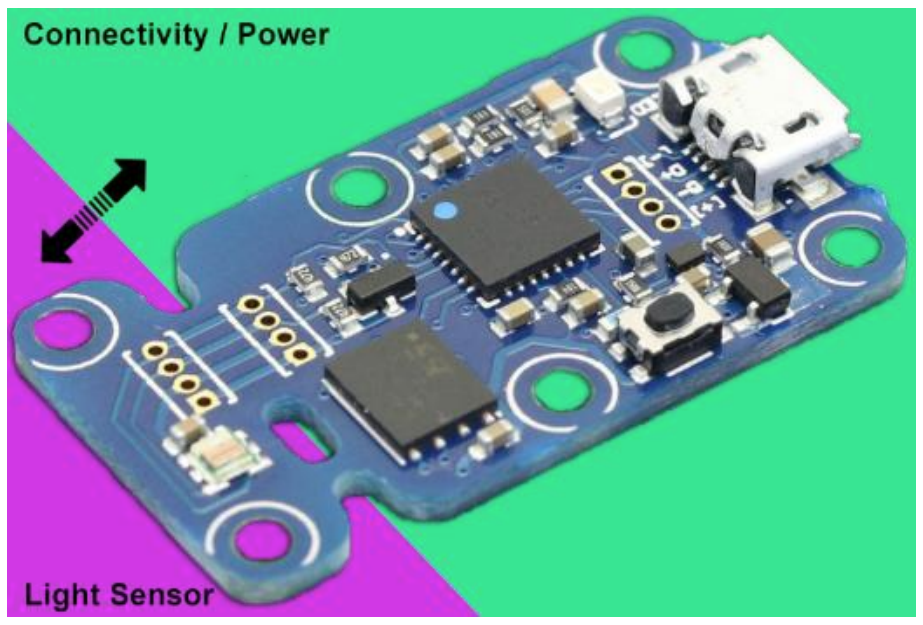

*Figure 8 Yocto-Light-V2 light sensor.*

*Adapted from* Yoctopuce Sarl, accessed 15 September, 2020, http://www.yoctopuce.com/projects/yoctolightV2/img/illustr-usb-light-sensor-1-big.jpg

A benefit of this light sensor is its size of only 20 x 35.5mm and a weight of 3g, which makes it well-suited for unobtrusive installations. For setups that do not provide enough space to mount the light sensor, the Yocto-Light-V2 has the option to separate the connectivity part (Figure 8, green area) from the actual sensor (Figure 8, magenta area). All that is needed is to solder a four-wire cable on the designated pads.

## Hardware Specifications

The hardware used for PESAO requires different parameters; hence the experimenter can alter possible independent variables to test the effects on dependent variables of the study. In the below table, we provide a list of hardware-parameters and their possible value range.

| Hardware | Parameter | Value Range |
|---|---|---|
| Eye-Tracking | Tracking Frequency | <50 / <100 Hz |
| | IMU Frequency | <50 / <100 Hz |
| | Gaze Tracking | <50 / <100 Hz |
| | Eye movement type | <50 / <100 Hz |
| | First-Person Camera | <25 FPS |
| Motion-Tracking | Tracking Frequency | <120 Hz |
| Light | Light Intensity | <7300 Lux/m |
| | Colour Temperature | 3200-5600 K |
| Camera | Frequency | <30 Hz |

*Table 1 PESAO Hardware Specifications.*

# Hardware Setup

After going over the necessary hardware for PESAO, we give a brief overview of how to put all pieces of hardware together. In addition, we also included a birds-eye camera to record subjects from above (Figure 9). In order to do this, almost any available webcam can be used.

Figure 9 gives an overview of how to integrate the hardware. Important to note is that PESAO relies on two different connectivity standards; USB 2.0 and WiFi ac. Components able to be connected over USB should be connected over a wired connection for robustness. However, the eye-tracking system, to allow unconstrained experiments, is connected over WiFi.

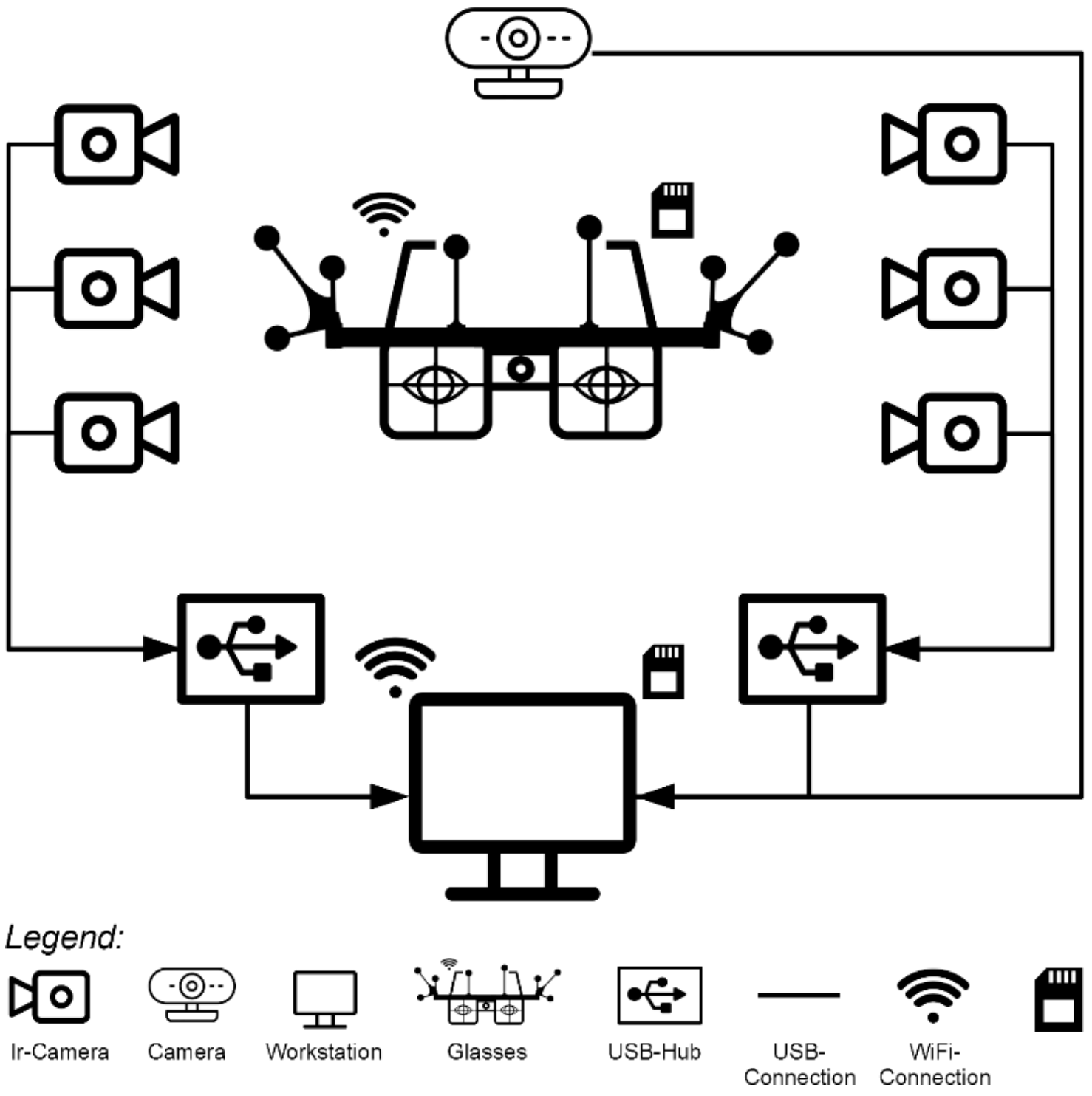

*Figure 9 PESAO connectivity overview.*

The WiFi connection is used to control and live view the eye-tracking system. Specifically, to set up, start and stop recordings and acquire synchronization timestamps of the system. The recording itself (gaze information, first-person video, calibration data) is stored on an SD-card in the recording unit of the eye-tracking system and will be copied after the experiment. Tobii includes with the *Tobii Pro Glasses 2 API* the capability of directly recording gaze data over WiFi, but in experiments, we have found a recording on the internal SD-card yields better tracking results.

Figure 10 shows an example setup of PESAO with dimensions. Important to note is that between the tracking and control area, a blackout curtain restricts the view so that the experimenter's movements do not distract the Subject.

<u>*Note:*</u> The Camera displayed in Figure 9 is in the current version of PESAOlib, not synchronized with the other sensors. However, this can be easily achieved as the video is stored as a MPEG file that carries a PTS (presentation timestamp) which can be used for synchronization.

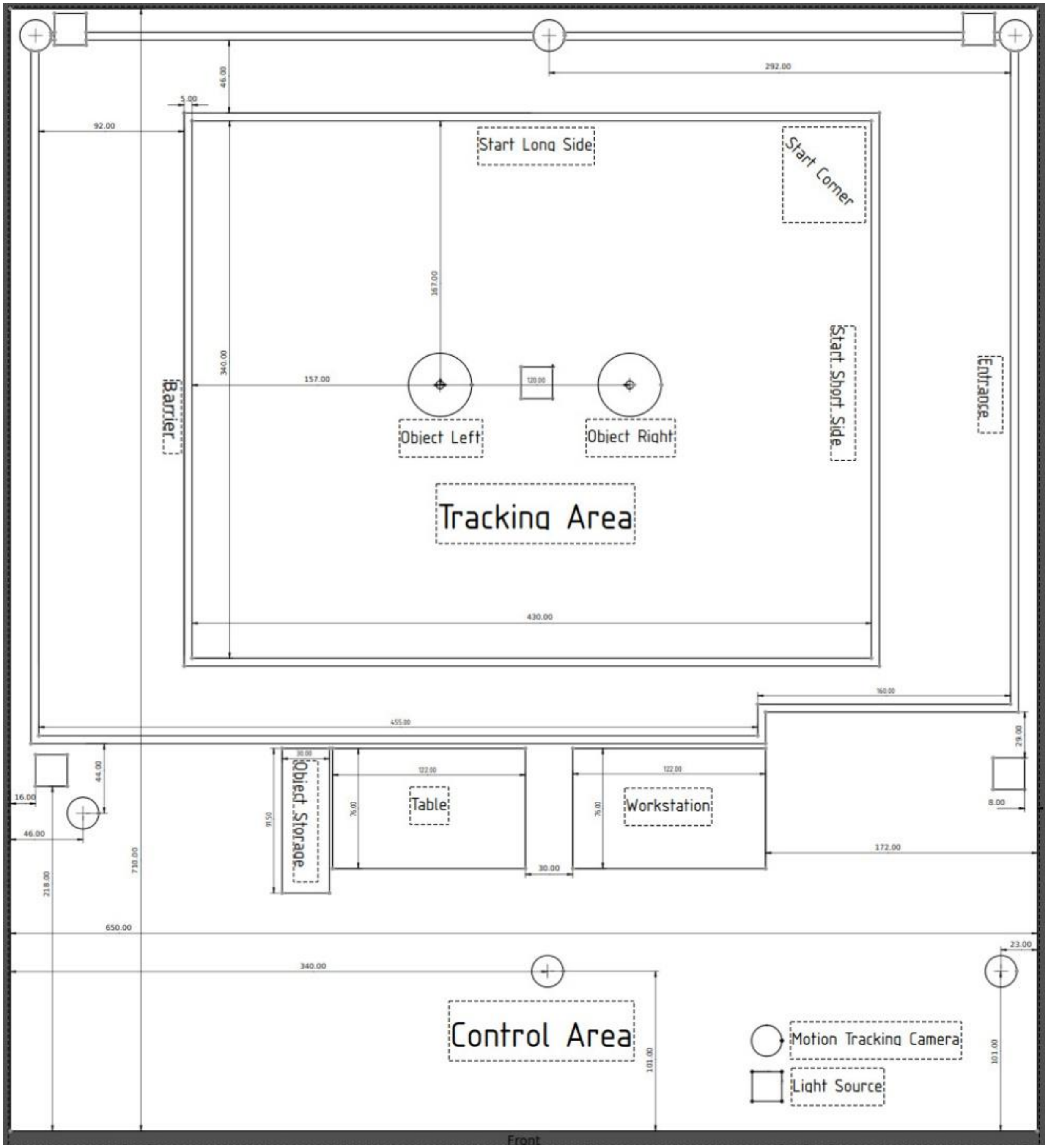

*Figure 10 PESAO example instantiation with dimensions. As used for this tech-report.*

# PESAOlib

PESAOlib is designed to control and execute experiments, record data with precise, accurate to microsecond-level timestamps, as well as synchronize and analyze the recorded data. PESAOlib provides a comma-separated values file (CSV) and a pickle file as output.

## PESAOlib overview

The software is written in Python and C++ and uses the networking and synchronization functionalities of the well-established *lab-streaming layer* by Swartz Center for Computational Neuroscience.

Provided are several source codes that are ready for compilation under Windows 10. Software parts that might vary from experimental design to another experimental design are provided in Python to be easier adjustable for programming novices.

Figure 11 displays a diagram showing the dependencies between PESAO's modules. All programs are executed on the Workstation (Figure 9). However, as PESAO is designed, programs can be run on multiple, connected workstations. This might be of interest if more sensors or a higher bandwidth is needed.

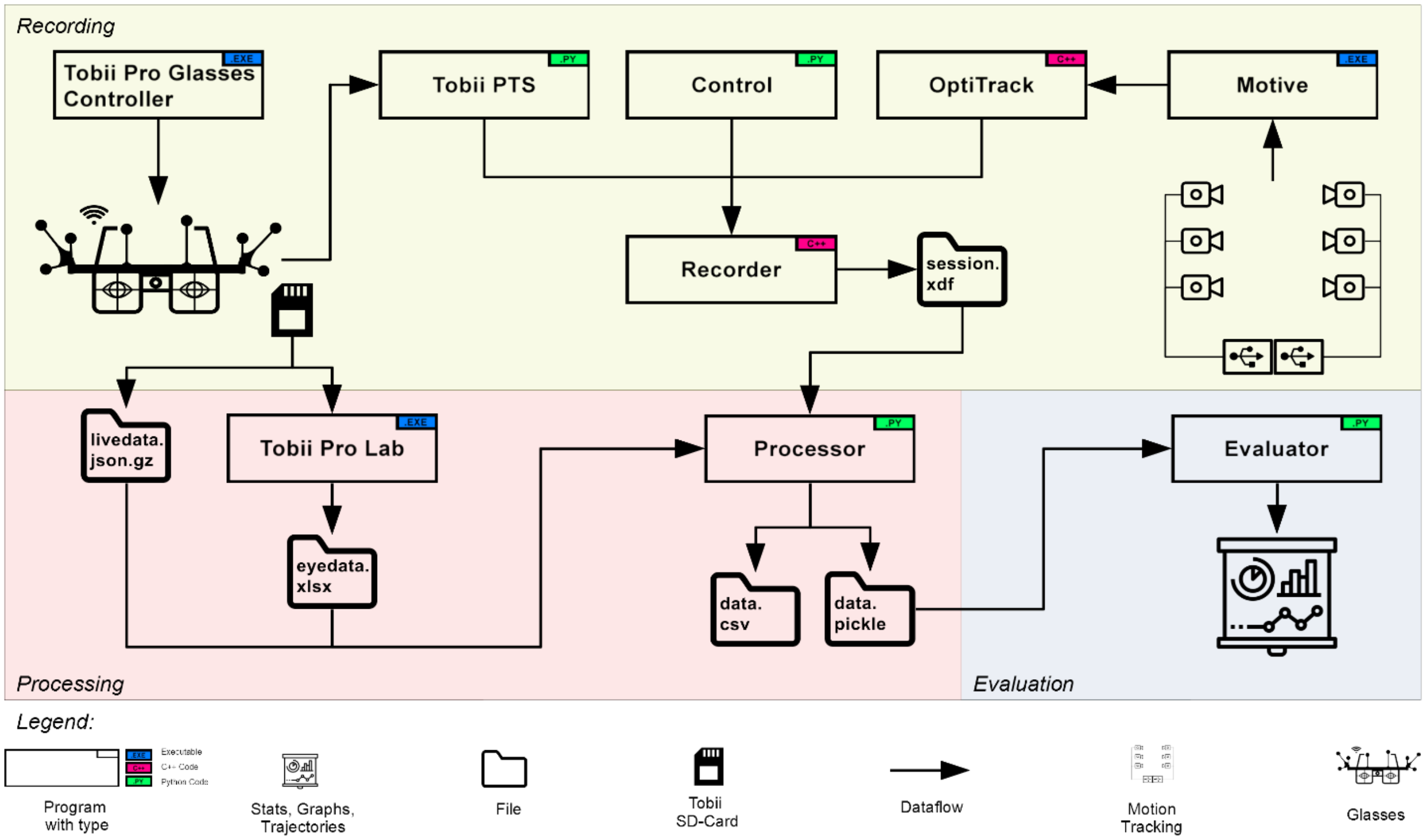

Figure 11 PESAOlib overview.

PESAOlib creates and saves files along its processing pipeline, to help the user to understand intermediate steps better and modularize their workflow. A strength of this system is that the Subject is entirely untethered, allowing free and natural motion. To compensate for transmission loss and lag, the eye-tracking device records on its local SD-Card.

As mentioned earlier, adjustable PEASAOlib modules are written in Python (green type indicator). However, modules that require high performance and are foreseen to remain unchanged for most setup scenarios are implemented in C++ (magenta type indicator). Nevertheless, the C++ source code is provided and can be changed as desired. Lastly, to allow the best interoperability with the physical devices, such as eye-tracking and motion-tracking system, we use the programs provided by the manufacturer (blue type indicator).

## PESAOlib Recording Module

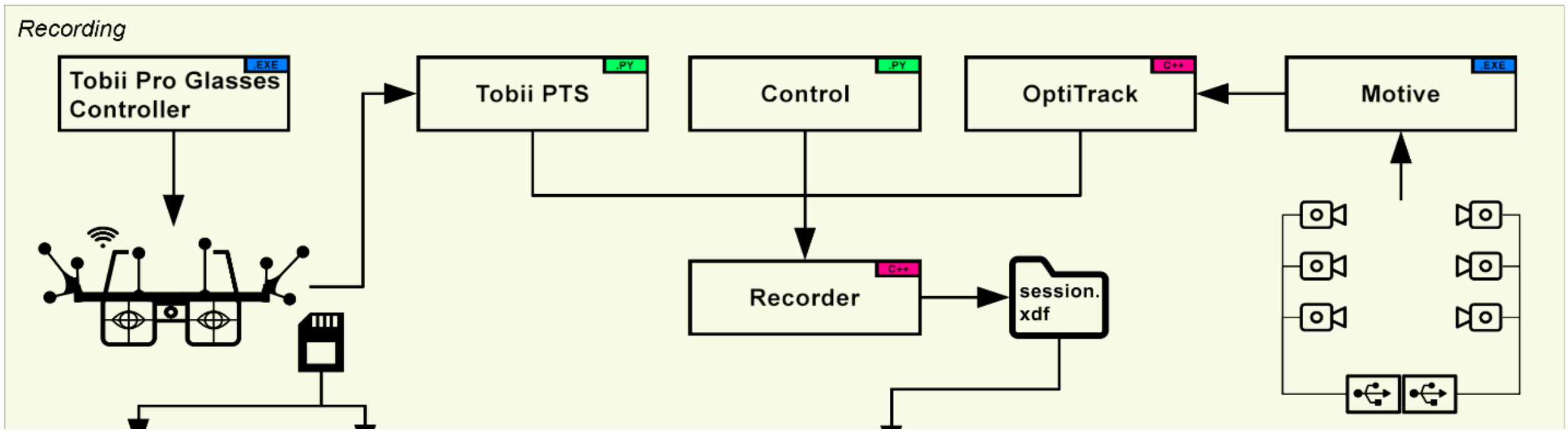

*Figure 12 PESAOlib Recording Section.*

Figure 11 top (bright green) depicts the *Recording* Section of PESAOlib. See Figure 12 for a close-up. These are modules responsible for gathering the data, including the Subject's gaze and head motion, synchronization timestamps, and experiment instructions, notes and feedback from the Subject.

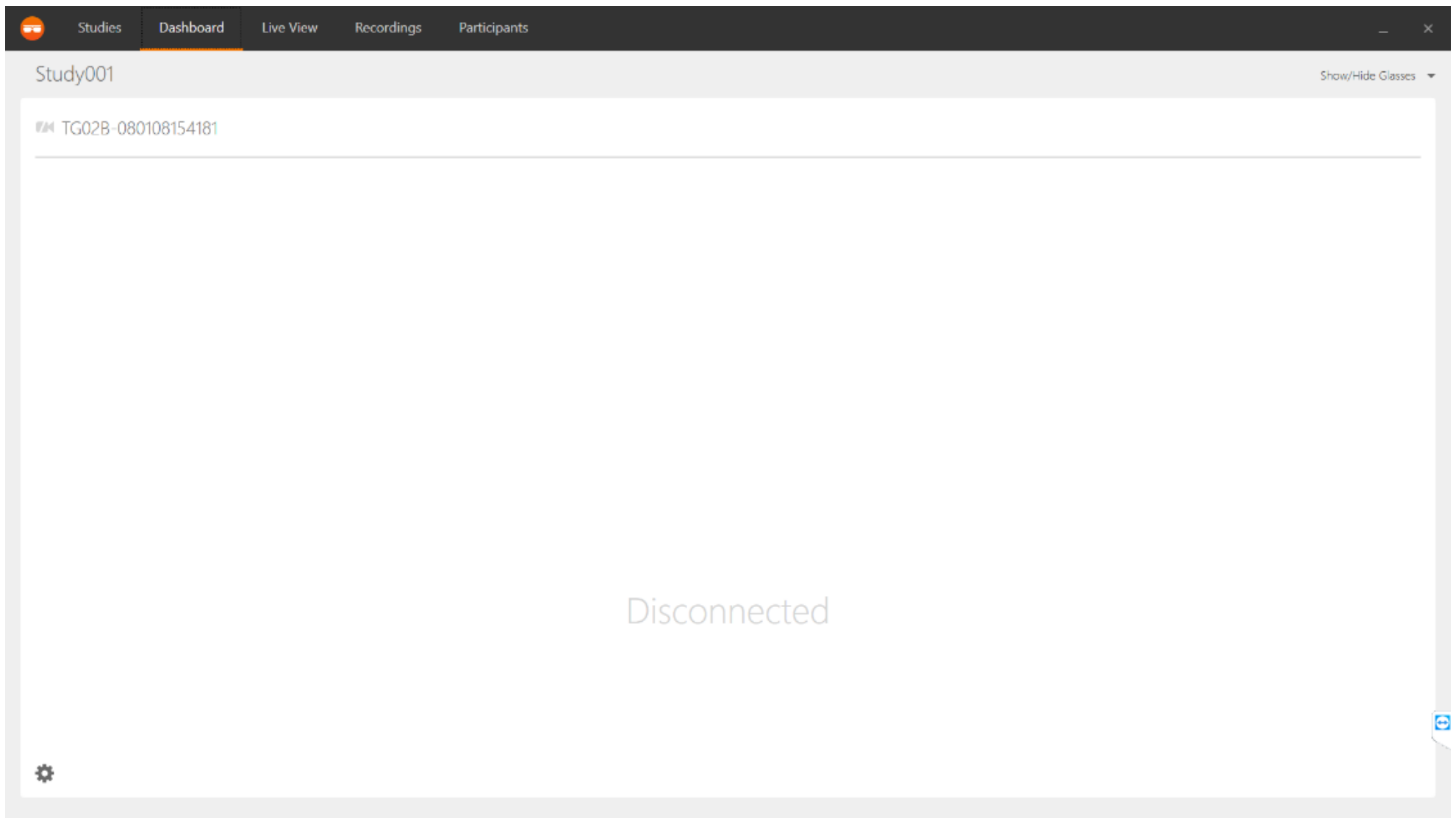

*Figure 13 Tobii Pro Glasses Controller.*

Starting on the top left, the executable *Tobii Pro Glasses Controller* controls, as the name suggests, the Tobii Pro Glasses II. With this program, you can set the Subject's name, check the battery and storage status of the glasses, calibrate the glasses, and start and stop recording. Figure 13 shows the dashboard of the program. It is straightforward to use. This includes an integrated update function to update the glasses' firmware.

*Tobii PTS* is a program written in Python that functions as a command-line program without a graphical user interface. Its task is to acquire recording timestamps from the glasses, which are used to synchronize the motion tracking system and the *control* program with the glasses' data. The program can be started at any point before the experiment, and it will begin collecting timestamps once the glasses are recording (initiated in *Tobii Pro Glasses Controller*).

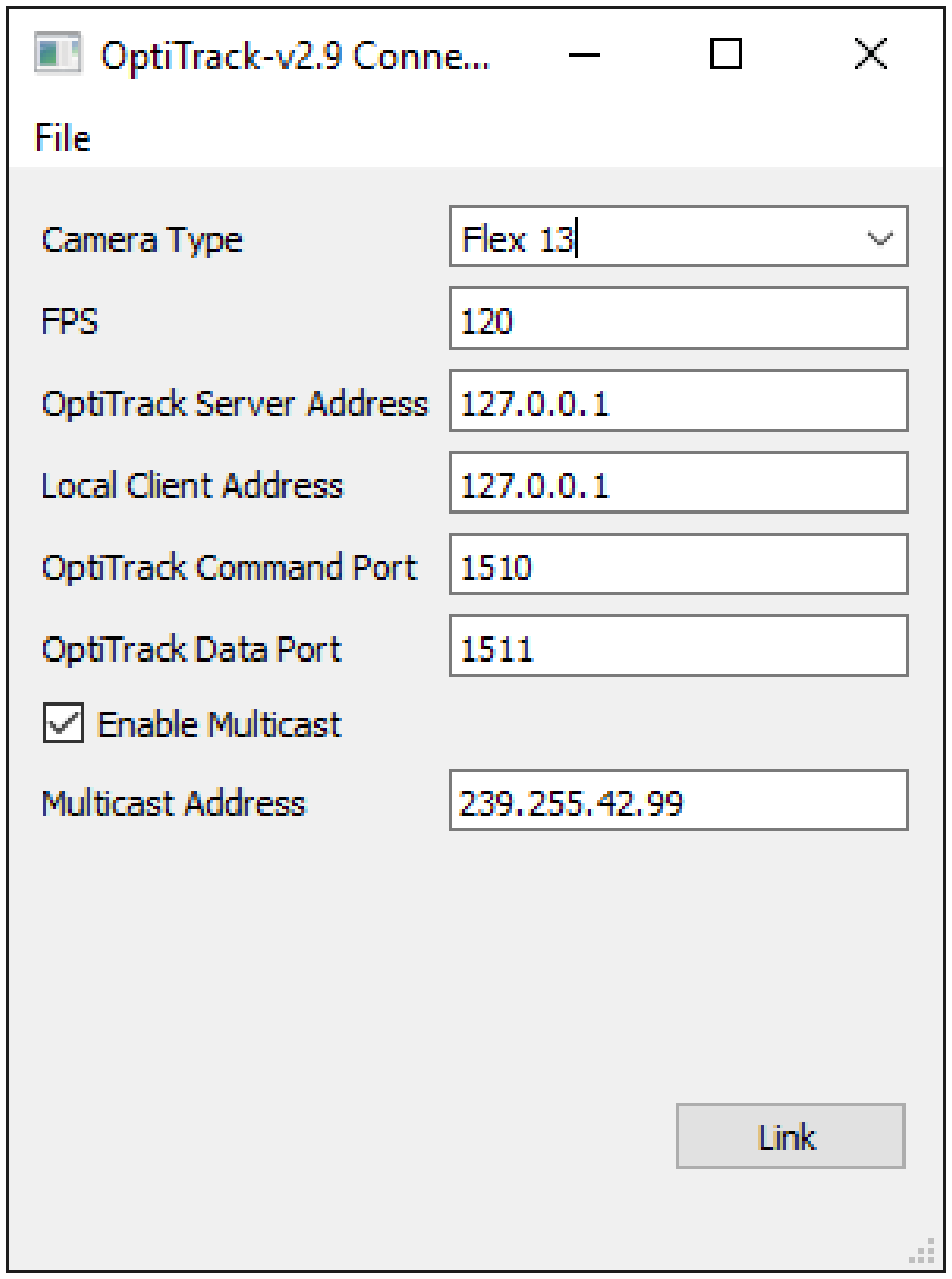

*Figure 14 PESAO OptiTrack Module.*

Similarly to the *Tobii PTS* program, *Control* is a Python command-line program. The task of this tool is to assist the experimenter with walking the Subject through the trials, generating randomized trials, taking observational notes, and recording the Subject's answers. This Python program is your starting point to design a new experiment.

For performance reasons, we implemented the program that gathers the live-streamed motion-tracking data in C++. With PESAO, we supply an executable that works with most OptiTrack cameras and also the source code. Figure 15 shows a screenshot of the user interface. The program also allows you to change the camera type, set frames per second, and a number of network settings to connect to *Motive*. However, the default network settings should suffice if *Motive* is run in default.

In order to calibrate and run motion tracking, PESAO relies on OptiTrack's *Motive* program. *Motive* is supplied with your OptiTrack Motion Tracking system — the program interfaces with tracking cameras and tracks the defined bodies. With PESAO, we supply a 3D head-model-file and 3D model-files of the object tracking bodies for visualization. Make sure to have *Motive's* live-streaming function enabled. Otherwise, OptiTrack will not be able to link to it. We recommend calibrating the motion tracking system at least once a day, and every time a tracking camera is moved. Figure 15 shows a screenshot of the user interface. Once the live-streaming functionalities have been enabled, all that is needed is to start *Motive*. It will automatically load the latest calibration, trackable bodies and starts live-streaming immediately.

<u>*Note:*</u> Do not press the red recording button. The *Recorder* program is our designated recording tool.

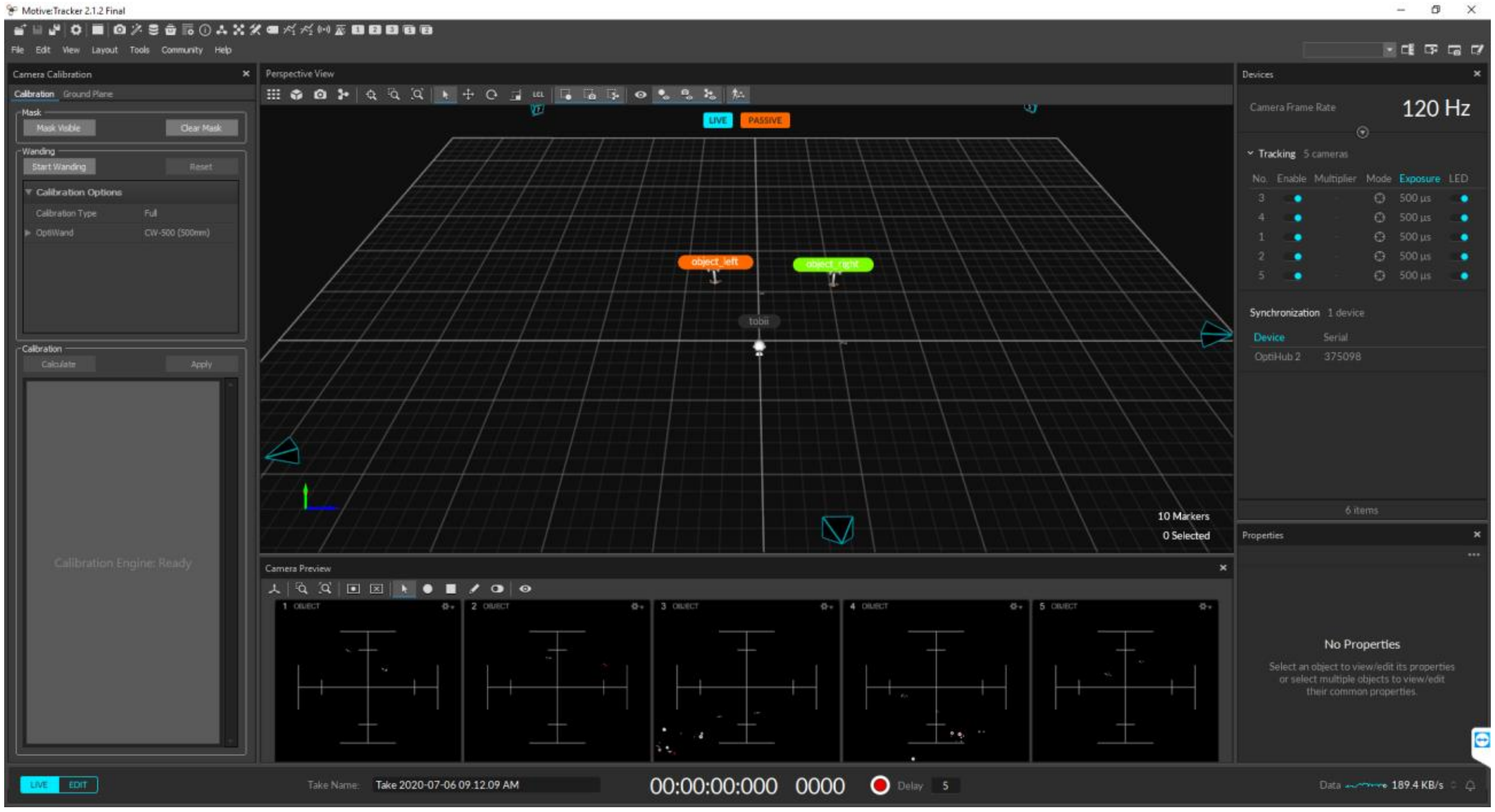

*Figure 15 OptiTrack's Motive user interface.*

The program that wraps everything together and is responsible for the recording is the *Recorder*. This program is implemented in C++ and comes with a user interface. See Figure 16.

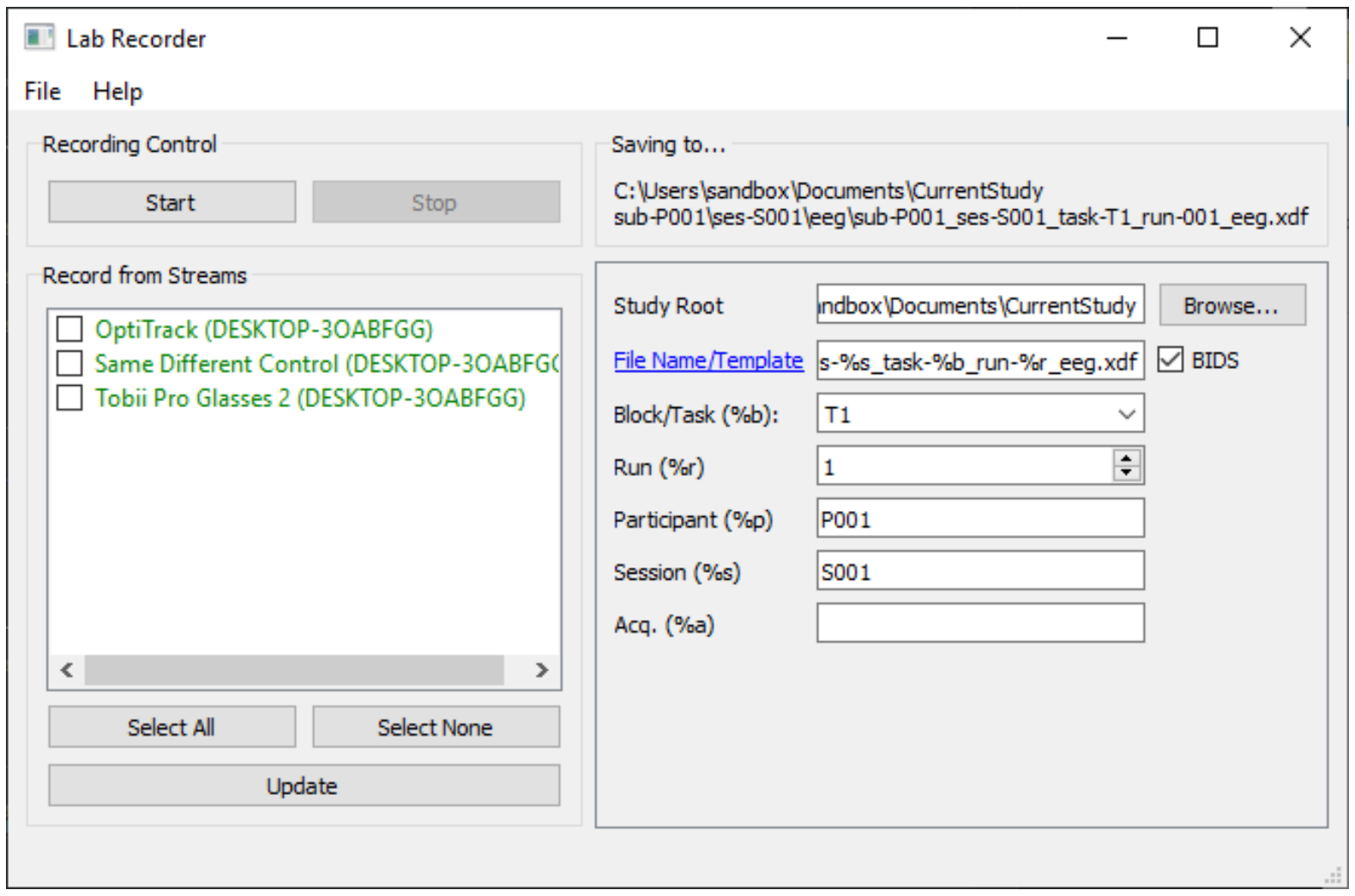

*Figure 16 PESAO Recorder.*

Following the startup of the previously mentioned programs, the *Recorder* is the last program needed to record an experiment. The user interface provides information about the available streams, and it should show three streams; *OptiTrack*, *Control*, and *Tobii PTS*. If the stream does not show and you are certain that all programs are running, press the Update button to search for available streams. On the right-hand side, you are able to set parameters for the experiment and Subject under investigation. The data will be saved into a XDF format in a folder specified under *Study Root*. To start recording, select all streams and press the *Start* button on the top left. To stop the recording press *Stop*. With this program, we conclude the *Recording* section of PESAOlib.

# PESAOlib Processing Module

After successfully recording the experiment, PESAOlib provides a processing program that cleans and synchronizes the motion tracking data with the control data and the gaze data. This Section is highlighted in a light red, and an enlarged version can be seen in Figure 17.

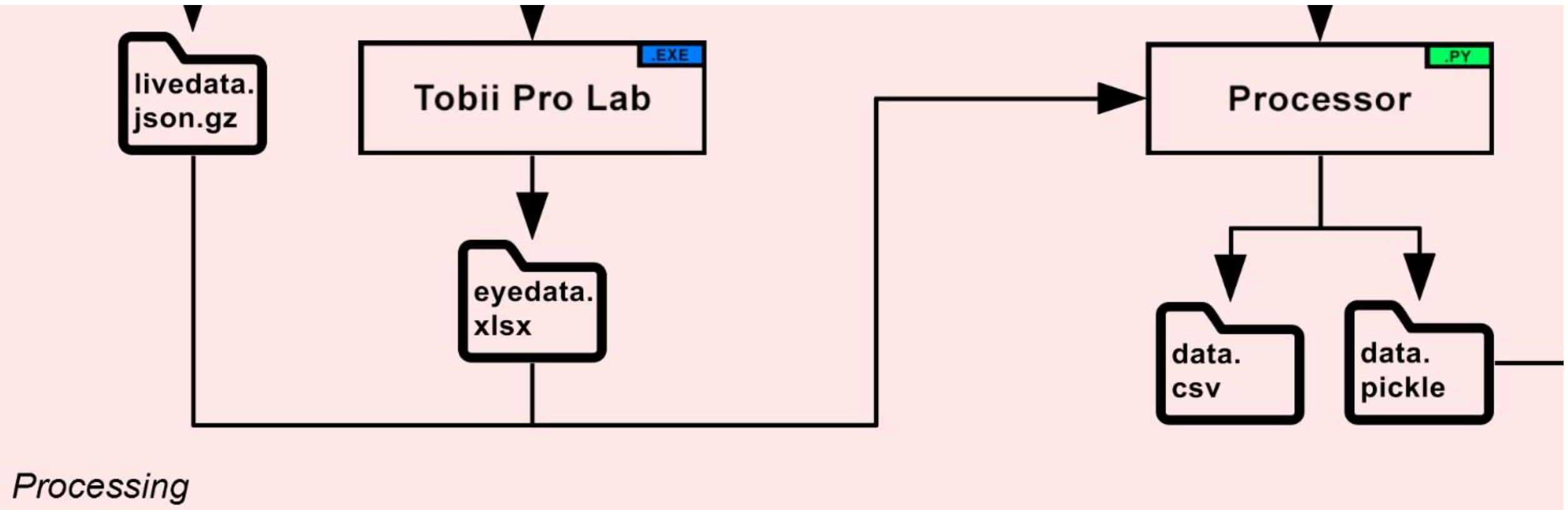

*Figure 17 PESAOlib Processing Section.*

As can be seen in Figure 11, the processor needs three files; the session.xdf produced by the *Recorder*, the livedata.json.gz, which can be loaded directly from the SD-Card of the glasses' recording unit and the eyedata.xlsx file. The eyedata.xlsx file has to be created by the analysis software *Tobii Pro Lab*. A screenshot of this software is displayed in Figure 18. Make sure to select the "Single Excel file (.xlsx)" as a setting for the format and select all data to be exported. The software provides more options to filter the eye-tracking data. These can be adjusted as needed.

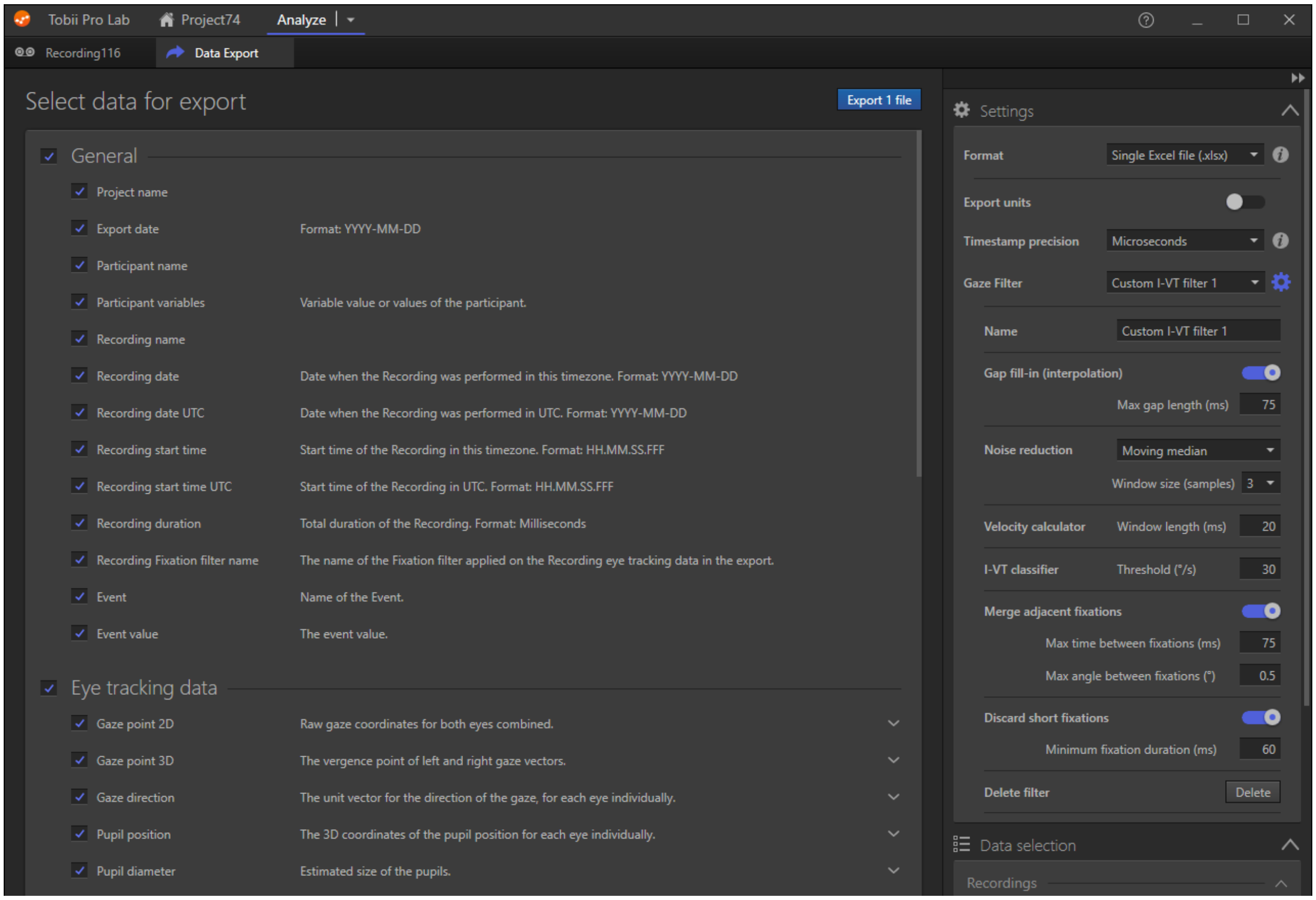

*Figure 18 Tobii Pro Lab export dialog. Showing the best settings for use with PESAO.*

## PESAOlib Evaluation Module

Lastly, PESAO provides with the PESAOlib *Evaluator* (Figure 11, light blue area) tools and examples to evaluate the generated data. PESAO was developed so that each process can be easily understood and examined. To realize this, besides providing source codes and developing PESAOlib mainly in Python, most of the generated artifacts are readable by humans (non-binary files) or a human-readable file is provided alongside the binary file (e.g. CSV file).

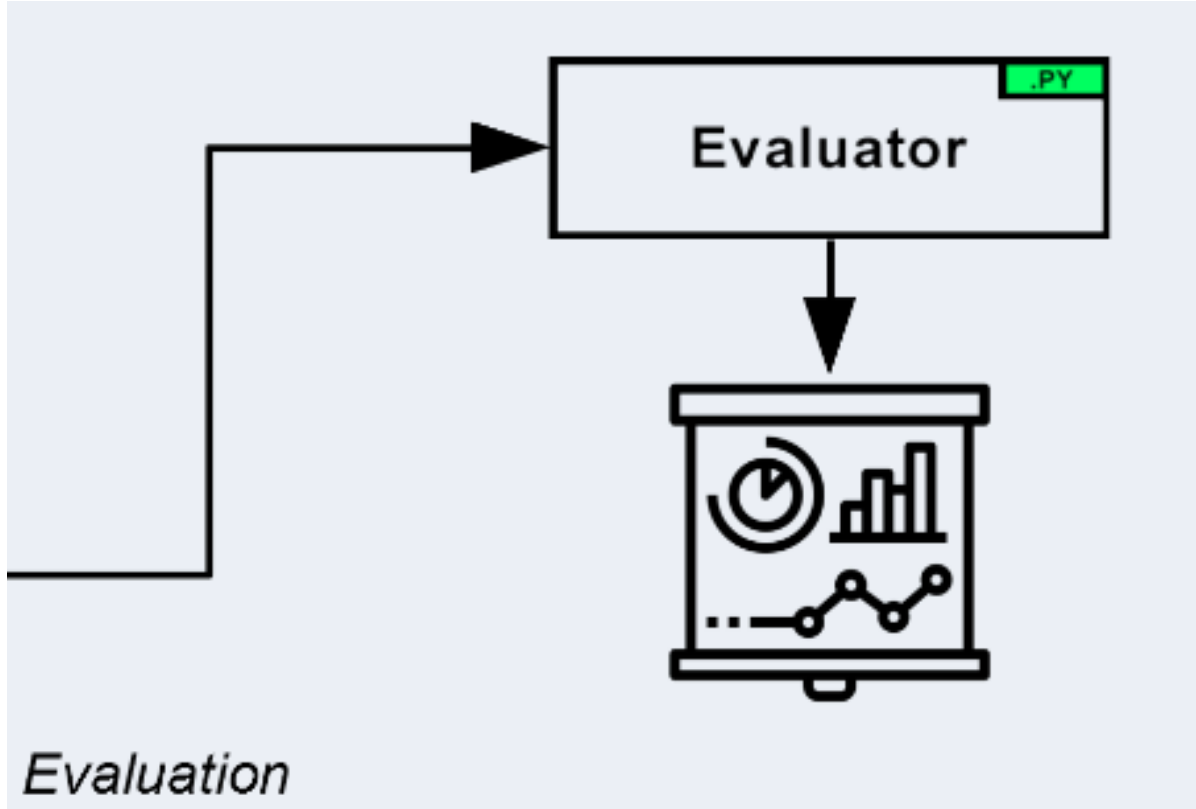

*Figure 19 PESAOlib Evaluation Section.*

The evaluation Section consists of the *Evaluator* program written in Python. The program supplies you with multiple utility classes to filter through *data.pickle*, plot graphs, annotate your plots with 3D STL models and more. Furthermore, the *Evaluator* program supplies you with a set of examples. One such example can be seen in Figure 20. It visualizes viewing frusta in temporal colour coding of subjects performing a visual task.

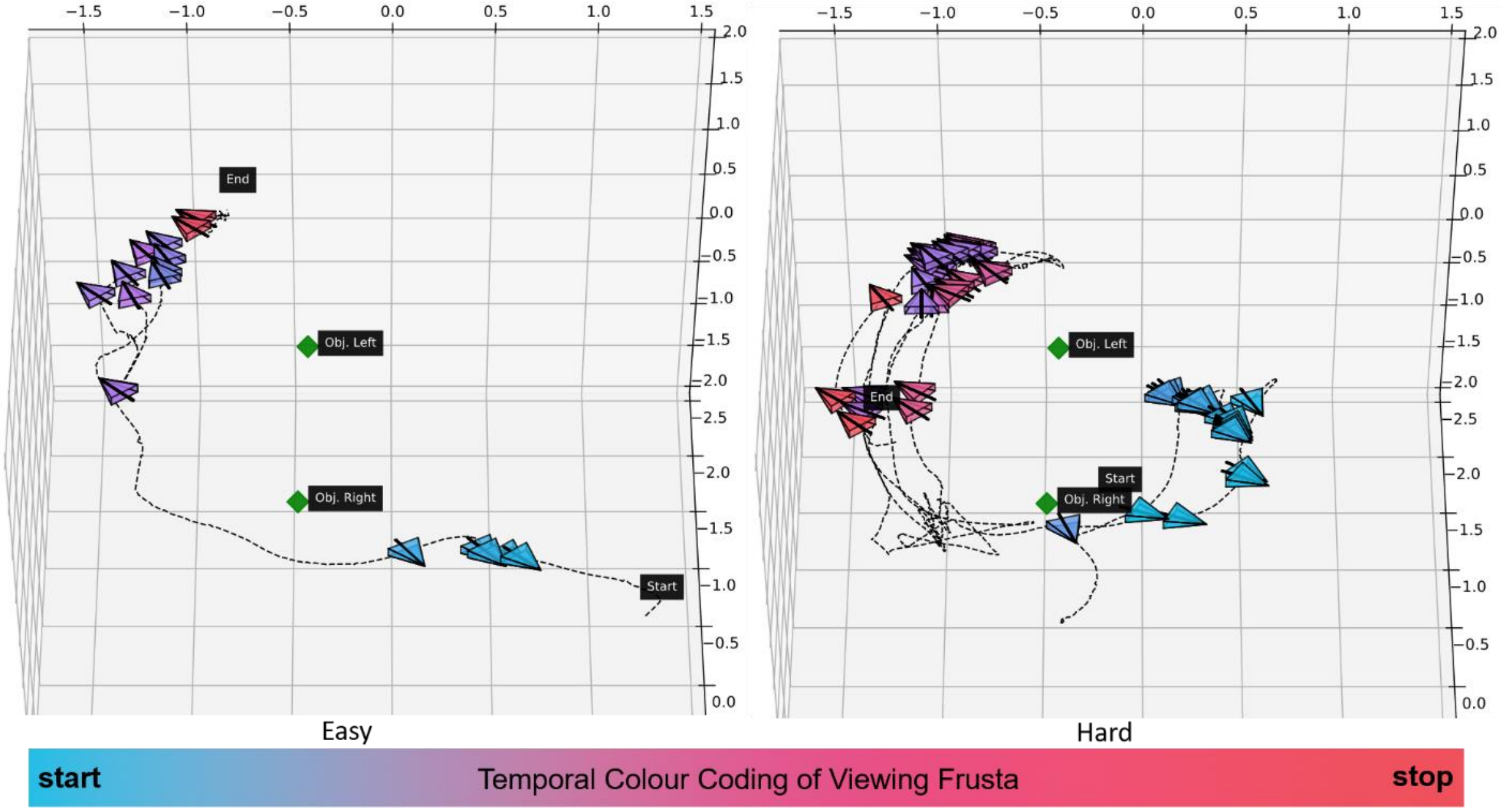

*Figure 20 Example visualization using PESAOlib Evaluator.*

## Project Page

In conclusion, PESAOlib provides you with three modules that cover everything from recording to processing and evaluation. Technical details can be found in the code documentation and check for updates for PESAOlib. PEASOlib is still in active development. To obtain a copy, please visit the GitLab home at https://gitlab.nvision2.eecs.yorku.ca/solbach/pesaolib/.

A screenshot of the project page of PESAO can be seen in Figure 21.

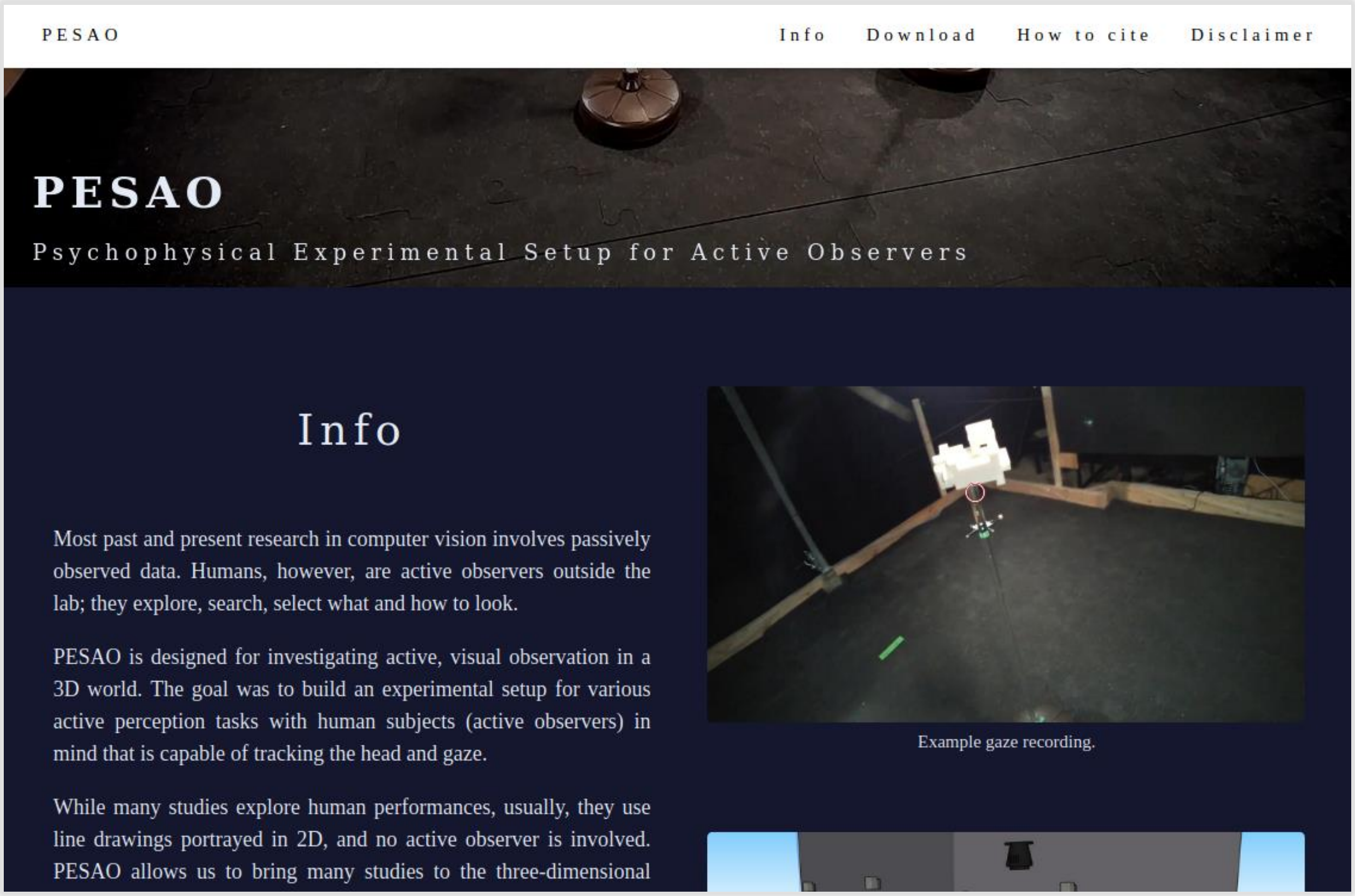

*Figure 21 Screenshot of the project page: http://data.nvision2.eecs.yorku.ca/PESAO/*

## Disclaimer

- We provide no guarantees for system performance.
- The system is provided as-is.
- Developed for research purposes only, and commercial use is not permitted.
- For comments, problems and questions, please send us an email (replies are not guaranteed).
- No level of technical support can be provided.
- If you use any aspect of this system, we respectfully request that all relevant publications that result from any use of this paper or system cite this paper.

## Citing this work

In case you use this work in one of your publications, please make sure to cite us.

```
Markus D. Solbach and John K. Tsotsos.
"PESAO: Psychophysical Experimental Setup for Active Observers"
arXiv preprint arXiv:2009.09933 (2020).
```